\documentclass{article}

\usepackage{microtype}
\usepackage{graphicx}
\usepackage{subcaption}
\usepackage{booktabs} % for professional tables

\usepackage{hyperref}

\usepackage[preprint]{icml2026}

\usepackage{amsmath}
\usepackage{amssymb}
\usepackage{mathtools}
\usepackage{amsthm}

\usepackage[capitalize,noabbrev]{cleveref}

\theoremstyle{plain}

\theoremstyle{definition}

\theoremstyle{remark}

\usepackage[textsize=tiny]{todonotes}
\usepackage{xcolor,color,soul}
\usepackage{tabularx}
\usepackage{enumitem}
\usepackage{xurl}

\colorlet{soulblue}{blue!20}

\usepackage{wrapfig}

\newlength\savewidth
\newcolumntype{x}[1]{>{\centering\arraybackslash}p{#1pt}}
\newcolumntype{y}[1]{>{\raggedright\arraybackslash}p{#1pt}}
\newcolumntype{z}[1]{>{\raggedleft\arraybackslash}p{#1pt}}

\definecolor{textgreen}{RGB}{57, 172, 57}
\definecolor{textred}{RGB}{200, 10, 10}
\usepackage{pifont} 
\newcommand{\cmark}{\textcolor{textgreen}{\ding{51}}}%
\newcommand{\xmark}{\textcolor{textred}{\ding{55}}}%

\icmltitlerunning{Wearable Foundation Models Should Go Beyond  Static Encoders}
\begin{document}

\twocolumn[
  % \icmltitle{Position: Foundation Models for Wearable Health Should Go Beyond Reactive, Snapshot-based and Pattern-recognition-centric Paradigms }
 %\icmltitle{Position: Wearable Foundation Models Should Go Beyond \\Pattern Recognition Paradigms }
  \icmltitle{Wearable Foundation Models Should Go Beyond Static Encoders}
  
  % It is OKAY to include author information, even for blind submissions: the
  % style file will automatically remove it for you unless you've provided
  % the [accepted] option to the icml2026 package.

  % List of affiliations: The first argument should be a (short) identifier you
  % will use later to specify author affiliations Academic affiliations
  % should list Department, University, City, Region, Country Industry
  % affiliations should list Company, City, Region, Country

  % You can specify symbols, otherwise they are numbered in order. Ideally, you
  % should not use this facility. Affiliations will be numbered in order of
  % appearance and this is the preferred way.
  \icmlsetsymbol{equal}{*}

  \begin{icmlauthorlist}
    \icmlauthor{Yu Yvonne Wu}{equal,dm}
    \icmlauthor{Yuwei Zhang}{equal,cam}
    \icmlauthor{Hyungjun Yoon}{equal,kai}
    \icmlauthor{Ting Dang}{mel}
    \icmlauthor{Dimitris Spathis}{goo,cam}
    \icmlauthor{Tong Xia}{qing}
    \icmlauthor{Qiang Yang}{cam}
    %\icmlauthor{}{sch}
    \icmlauthor{Jing Han}{cam}
    \icmlauthor{Dong Ma}{cam}
    \icmlauthor{Sung-Ju Lee}{kai}
    \icmlauthor{Cecilia Mascolo}{cam}
    %\icmlauthor{}{sch}
    %\icmlauthor{}{sch}
  \end{icmlauthorlist}
  % \printAffiliationsAndNotice{\icmlEqualContribution}
  % \printAffiliationsAndNotice{${}^*$Equal contribution}

  \icmlaffiliation{dm}{Dartmouth College, USA}
  \icmlaffiliation{cam}{University of Cambridge, UK}
  \icmlaffiliation{kai}{KAIST, South Korea}
  \icmlaffiliation{mel}{The University of Melbourne, Australia}
  \icmlaffiliation{goo}{Google Research, UK}
  \icmlaffiliation{qing}{Tsinghua University, China}

  \icmlcorrespondingauthor{Yu Yvonne Wu}{yvonne.wu@dartmouth.edu}
  \icmlcorrespondingauthor{Yuwei Zhang}{yz798@cam.ac.uk}
  \icmlcorrespondingauthor{Hyungjun Yoon}{hyungjun.yoon@kaist.ac.kr}

  % You may provide any keywords that you find helpful for describing your
  % paper; these are used to populate the "keywords" metadata in the PDF but
  % will not be shown in the document
  \icmlkeywords{Machine Learning, ICML}

  \vskip 0.3in
]

% this must go after the closing bracket ] following \twocolumn[ ...

% This command actually creates the footnote in the first column listing the
% affiliations and the copyright notice. The command takes one argument, which
% is text to display at the start of the footnote. The \icmlEqualContribution
% command is standard text for equal contribution. Remove it (just {}) if you
% do not need this facility.

% Use ONE of the following lines. DO NOT remove the command.
% If you have no special notice, KEEP empty braces:
% \printAffiliationsAndNotice{}  % no special notice (required even if empty)
% Or, if applicable, use the standard equal contribution text:
\printAffiliationsAndNotice{\icmlEqualContribution}

\begin{abstract}
% Wearable health monitoring, enabled by daily-affordable devices that continuously capture our physiological and behavioral data, has become increasingly prevalent.
% Together with foundation models leveraging vast amount of unlabeled data, wearable foundation models (WFMs) have shown strong performance on benchmark tasks such as human activity recognition and sleep staging. 
% \ds{Re: title, I would argue that all the proposed solutions (longitudinal modeling, agentic inference, trajectory forecasting) are still pattern recognition.. What if we focus on the fact that existing models are just encoders? Title suggestion: ... "should go beyond static encoders"} \yz{I don't mind the suggestion?@yvonne@Jun} \yz{conclusion: we like this but do not have enough time to reframe across the entire paper.}
Wearable foundation models~(WFMs), trained on large volumes of data collected by affordable, always-on devices, have demonstrated strong performance on short-term, well-defined health monitoring tasks, including activity recognition, fitness tracking, and cardiovascular signal assessment.
% \cm{i wonder if we could give more ambitious examples than vs monitoring, eg cardiovascular health monitoring}. 
However, most existing WFMs primarily map short temporal windows to predefined labels via static encoders, emphasizing retrospective prediction rather than reasoning over evolving personal history, context, and future risk trajectories. As a result, they are poorly suited for modeling chronic, progressive, or episodic health conditions that unfold over weeks, months or years. %This emphasis limits their capacity to capture and interpret more complex, longitudinal health conditions. 
%For this reason, in
% \cm{and why is this important? perhaps we can say that without these, more advanced understanding of our health is limited} In 
Hence, \textit{we argue that 
% \cm{add for this reason (in case you add a sentence above too) } 
WFMs must move beyond 
% short-horizon\hj{how about removing short-horizon from both here and the title? it also makes the abstract to be within a single column} \ds{agree }
static encoders and be explicitly designed for longitudinal, anticipatory health reasoning}. We identify three foundational shifts required to enable this transition:
%dimensions that should shift to support more anticipatory and human-aligned health monitoring:
(1)~\textit{Structurally rich data},
% \cm{unclear if you mean just that we should have more data or being able to model more aspects ie multimodal},
which goes beyond isolated datasets or outcome-conditioned collection to integrated multimodal, long-term personal trajectories, and contextual metadata, ideally supported by open and interoperable data ecosystems;
%structurally rich wearable data, combining multimodal sensor signals, longitudinal trajectories, and contextual information within open data ecosystems;
(2)~\textit{Longitudinal-aware multimodal modeling}, which prioritizes long-context inference, temporal abstraction, and personalization over cross-sectional or population-level prediction; and
% \qy{Seems like there is a partial overlap between 1 and 2: which one emphasizes multimodal?}
(3)~\textit{Agentic inference systems}, which move beyond static prediction to support planning, decision-making, and clinically grounded intervention under uncertainty. 
Together, these shifts reframe wearable health monitoring from retrospective signal interpretation toward continuous, anticipatory, and human-aligned health support. \
\end{abstract}

\section{Introduction}

\begin{figure*}
    \centering
    \includegraphics[width=\linewidth]{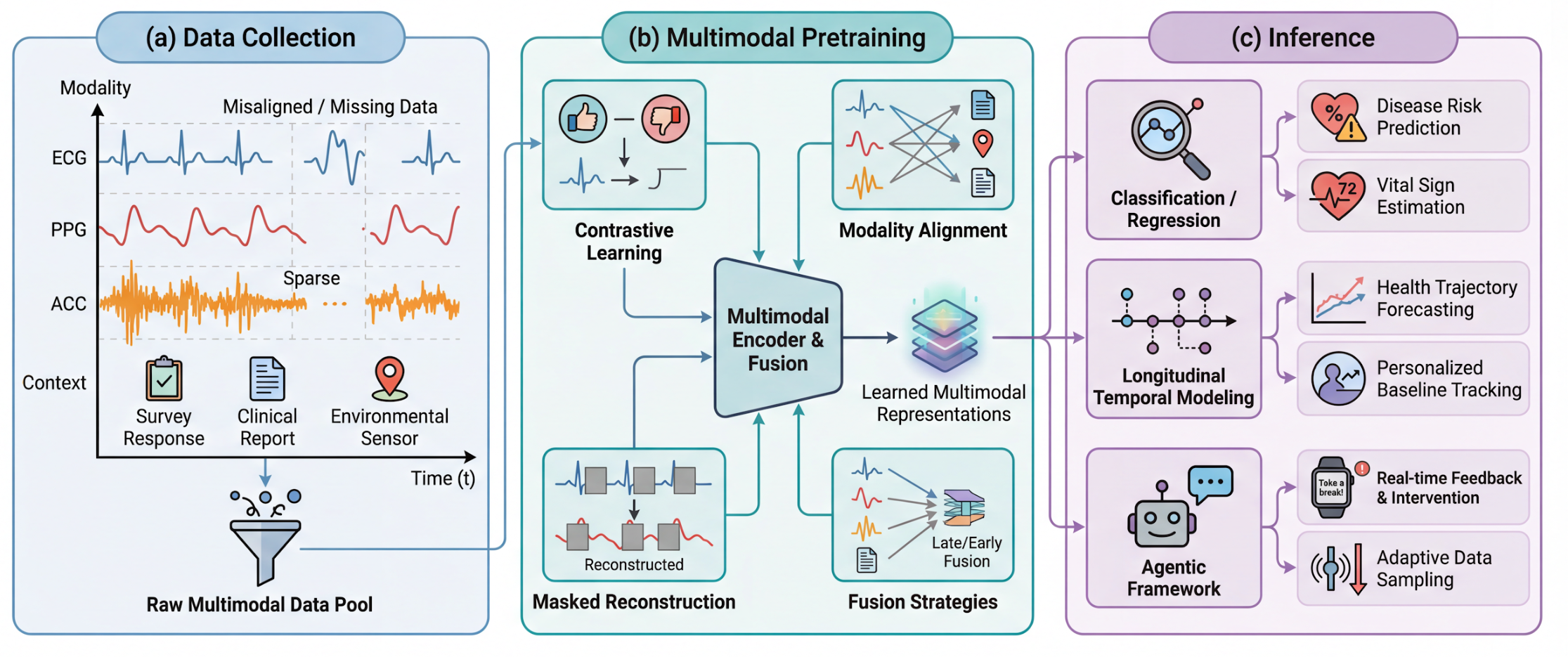}
    \caption{\textbf{Overview of a Wearable Foundation Model}. The pipeline consists of three main stages: (a) Data Collection, involving heterogeneous physiological signals and contextual metadata; (b) Multimodal Pretraining, which utilizes self-supervised learning and cross-modality alignment to produce robust latent representations; and (c) Inference, where the model is applied to downstream tasks.}
    % \yz{placeholder, don't mind the redundant arrows}. \qy{This figure could be moved to the first page.}}
    \label{fig:overview}
    % \vspace{-0.3cm}
\end{figure*}

% Today’s healthcare system is largely reactive and faces well-known bottlenecks, including resource scarcity and long wait times. \hj{Can we say ``Today's''? I think we are claiming the problems of foundation models, which contradicts with this statement.} To improve and complement this system, the AI community has pursued two broad directions. One focuses on clinical automation (\textit{e.g.}, Med-Gemma, Med-ASR), aiming to optimize physicians’ workflows within clinical settings. Meanwhile, A second, increasingly important direction seeks to extend health support beyond the clinic by leveraging digital and sensor-based technologies for everyday health monitoring~\citep{}.
% \td{As wearable foundation models generally target applications outside clinical settings, broader healthcare system management is beyond their scope. It may be clearer to start the discussion directly with wearable foundation models or the relevant scopes.}

% In particular, wearable foundation models (WFMs) represent a particularly promising yet underexplored opportunity. \hj{imo, we can start from this paragraph. with few more sentences - e.g., AI community enabled ... in wearable health monitoring ...}
Recent advances in wearable sensing have enabled a new class of machine learning systems for continuous, everyday health monitoring.
Modern wearables, including smartwatches and fitness trackers, continuously collect multimodal physiological and behavioral signals, such as cardiovascular activity (ECG/PPG), motion (IMU), respiration, and ambient temperature, often over months or years. 
When paired with foundation models trained on large-scale data, these systems, which we refer to as \textit{Wearable Foundation Models~(WFMs)}, have demonstrated strong analytical performance on a range of well-scoped analytical tasks, including human activity recognition, sleep staging, and population-level health monitoring~\citep{Narayanswamy2025lsm, thapa2026sleepfm}. For example, SensorLM~\citep{zhang2025sensorlm}, trained on population-scale Fitbit data, exhibits strong zero-shot generalization for activity recognition. In parallel, large language model~(LLM)-based health-assistants, such as ChatGPT Health\footnote{https://openai.com/index/introducing-chatgpt-health} and Claude Healthcare\footnote{https://www.anthropic.com/news/healthcare-life-sciences}, have begun  integrating wearable signals into conversational systems intended to provide health-related guidance when clinicians are unavailable.
% \yz{removed}
% , \hj{the following sentence does not fit well with this paragraph: challenges should be in the next paragraph}\td{just remove?}but are often limited to text-based interactions and sparse integration of real-world health data. WFMs represent a new interface between machine learning systems and human health, one that operates continuously, longitudinally, and outside the clinic.\cm{this last part does not seem to follow the flow well}

%%%%%%%should we still talk about what the community has achieved so far??

%%% this part motivate why current ML for wearable healcare are insufficient
Despite this progress, most existing WFMs remain fundamentally confined to
% grounded\cm{grounded does not mean limi ted, i would say confined to} in 
% short-horizon encoders and result in static pattern recognition paradigms
static encoders, which frame wearable health modeling as a form of static pattern recognition over short temporal windows~\citep{abbaspourazad2024largescaletrainingfoundationmodels}. Specifically, they map short segments of physiological and behavioral data to predefined labels or scalar targets, such as activity classes, sleep stages, or fitness scores.
Such a static paradigm  closely mirrors paradigms that have proven effective in vision and speech recognition, where inputs are typically treated as independent, stationary, and semantically complete within short temporal windows~\citep{luo2024normwear, thapa2026sleepfm}. 
% \cm{here i would say that if we only have this type of mapping we cannot do thee things that we need to be able to do with these systems ie long term health tracking etc. the sentence below is just talking about the signal instead of the final goal}
% Wearable health monitoring differ fundamentally: signals are continuous, longitudinal, and context-dependent, and meanwhile clinical and behavioral meaning often emerging only over extended histories and changing conditions (over days, months and even years)
% \yz{todo:add more about the goal here}. 
However, when applied to wearable health, this framing directly limits the system's ability to support  %as a mapping from short-term sensor segments to predefined targets precludes 
core health objectives, including longitudinal health tracking, personalized baseline estimation, and reasoning about gradual change, recovery, or disease progression.
% These limitations are systematically present across the WFM pipeline, spanning data, modeling, and inference.
Importantly, these limitations are not isolated deficiencies of individual models or datasets; they are systematic consequences of a shared static encoder framing that permeates the entire WFM pipeline, spanning data collection, modeling, and inference.

%%%data-wise, --> 
%%%%%should we conclude it as re-active problem???
At the data level, the dominance of such static paradigm is reflected in how most wearable datasets are constructed. Most open-source wearable datasets are organized around
% constructed\cm{repetition of word} around 
short temporal windows, isolated sensing modalities, and outcome-conditioned labeling schemes, 
% with data  collected or curated to support 
in support of narrowly defined downstream tasks~\citep{stisen2015hhar, kemp2000sleepedf}. 
% Such designs implicitly assume that health-relevant information is localized, stationary, and semantically complete within brief signal segments, an assumption that aligns well with pattern recognition but poorly with real-world health monitoring.
Although large volumes of passive sensor data may be collected, richer contextual information, such as behavior, environment, device usage, or self-reported state, is often sparse, inconsistently integrated, or entirely unavailable. Such contextual factors are critical for causal and situational grounding, enabling interpretation of \textit{why} physiological patterns arise rather than merely \textit{what} patterns are present~\citep{abowd1999context}. As a result, WFMs trained on these datasets are structurally biased toward learning correlations, rather than reasoning about evolving, personalized health states over time.
Moreover, many population-scale multimodal and longitudinal wearable datasets remain proprietary, limiting reproducibility, benchmarking, and cross-dataset evaluation~\citep{Narayanswamy2025lsm, erturk2025beyond}. Consequently, existing wearable datasets largely support WFMs that function as %small-scale 
passive observers, optimized for short-term pattern recognition rather than longitudinal health reasoning.
% \cm{i am not sure i would put this as a first point as there is very little we can suggest a part from pointing at efforst such as biobank}
% rather systems capable of to reason about evolving health trajectories from long history and environmental context. 

%%modeling, mainly two challenges?? do not have 
At the modeling level, the static paradigm is reinforced by how WFMs represent
multimodal and temporal information. Most dominant approaches inherit these dataset-level assumptions by treating inputs as independent temporal segments and often training separate foundation models for individual sensing modalities~\citep{gu2025foundation}.
More recent work has explored foundation models spanning multiple sensor modalities and proposed a range of multimodal pretraining strategies~\citep{Narayanswamy2025lsm, luo2024normwear, erturk2025beyond}. However, these methods remain architecturally tied to fixed set of input modalities and temporal resolutions, typically relying on data- and task-specific multimodal fusion mechanisms. As a result, existing WFMs tend to internalize modality- and resolution-specific patterns during pretraining, making them difficult to generalize to heterogeneous sensor configurations, missing modalities, or variable sequence lengths encountered in real-world deployments. 
% they do not generalize to heterogeneous sensor configurations or input sequnece length, limiting their real-world usability
% \td{reads more like data perspective?rephrase a bit?}. 
These architectures also struggle to natively incorporate contextual information, often neglecting it altogether or deferring it to ad hoc late fusion with language models.
% \hj{I think context is also covered in the dataset part.}
% However, they differ in the modalities covered, tokenization used, and are still far from enabling arbitrary set of modalities as input. 
More fundamentally, the dominance of short-window, cross-sectional modeling prevents WFMs from effectively leveraging longitudinal health trajectories, undermining the primary advantage of wearable sensing: ubiquitous, long-term, everyday monitoring. 
% \qy{You also mentioned longitudinal design in the next paragraph.}

% , or relying on data and task-specific
% % fragmented \yz{how can fusion methods be fragmented...do we mean non-standardized? data and method-specific?} 
% multimodal fusion strategies. 
% Even when multimodal signals are available, 
% existing work spans a wide range of modeling strategies, reflecting the absence of a common abstraction for multimodal physiological modeling

%%inference-level.. Models output predictions, not decisions No notion of: planning, adaptive sensing ..Systems respond after patterns emerge, not while trajectories shift
Furthermore, at the inference level, most existing WFMs remain confined to functioning as discriminative predictors, mapping incoming signals to predefined labels or scores, rather than supporting agentic inference that reasons adaptively over evolving observations and history~\citep{yang2023autogpt}. 
% \cm{and what would this enable?} \qy{I guess you are referring to proactive/adaptive systems which can be achieved by the agentic design}
Under this framing, health monitoring is treated as a static prediction problem, in which outputs such as risk scores or activity classes are generated independently at each time step.
While effective for benchmarked prediction tasks, this formulation provides limited support for real-world health scenarios that require more than pattern detection, specifically, reasoning about temporal evolution, accumulating evidence across observations, and determining when additional sensing, feedback, or intervention is warranted~\citep{yang2025contextagent}.
% As a result, current systems primarily react to salient patterns after they emerge, rather than acting as agentic systems that can guide what to monitor, how to interpret accumulating signals, and when additional evidence is needed. \qy{Agent is one of the ways to do it} 
As a result, this predictive-only formulation constrains downstream capabilities such as adaptive intervention planning, personalized behavioral recommendations, and longitudinal decision support, thereby motivating a shift from static prediction toward agentic inference in wearable health systems.
% This gap between predictive modeling and real-world health inference motivates a shift from discriminative prediction toward agentic inference. 

% \qy{After reading this, I feel now these three pillars are entangled, e.g., the fragmented data window/longitudinal/multimodal issues are introduced across two/three subsections. Not sure if we need to make it more structural}

% \textbf{Our position:} 

We argue that the dominant limitation of existing WFMs is not model capacity, but their shared reliance on a static encoder that frames wearable health as the retrospective mapping of short-term sensor windows to predefined targets.
The next generation of foundation models for wearable health must move beyond short-horizon static encoders and be explicitly designed for longitudinal, anticipatory health reasoning.
% Achieving this requires rethinking (1) how data are collected and represented, (2) how multimodal and longitudinal information is modeled, and (3) how model outputs are translated into actionable, human-aligned health support.
Moving beyond static encoders requires coordinated shifts across the full WFM pipeline, spanning data, modeling, and inference:
\begin{itemize}[nosep, leftmargin=*]
    % \item  \textbf{Comprehensive Analysis of Prior Work and Alternative Viewpoints:} XXXXX  
    % \item \textbf{Identification of Key Challenges and Gaps:} XXXXXX  
    % \item \textbf{Articulate foundation guidelines}: 1) data richness; 2) longitudinal-aware multimodal modeling 3)proactive agentic systems
    \item \textbf{Structurally rich data}, emphasizing open and interoperable data ecosystems that integrate %containing structurally rich wearable data, combining 
    multimodal sensing, long-term personal trajectories, and contextual information beyond raw signals. 
    % emphasizing open, heterogeneous, and longitudinal data ecosystems that integrate multimodal and contextual signals
    % \jh{not just signals but information? some data is not signal level?}. \cm{I would mention examples of efforts trying to collect them eg biobank}
    \item \textbf{Longitudinal-aware multimodal modeling}, enabling long-context inference and  personalization over multimodal health trajectories rather than isolated observations.
    \item \textbf{Agentic inference}, transforming model outputs into adaptive, decision-oriented, and human-aligned health support rather than static predictions.
\end{itemize}
\section{Overview of Wearable Foundation Models}

Wearable Foundation Models (WFMs) operate at the intersection of continuous sensing, large-scale representation learning, and real-world health decision support. In this section, we provide a unifying overview of the WFMs pipeline, illustrated in Figure~\ref{fig:overview}. We highlight the core design considerations across the key pillars of a WFM: data acquisition, multimodal pretraining, and inference for health support.
% \qy{I suggest we add more explanations to this overview as few people will read details (from reviewers' view)} \yz{I think better scope here that we focus on multimodal. otherwise both dataset and model table is incomplete.} 

Given that wearable data is inherently multimodal, we analyze each pipeline stage through a multimodal lens. We treat unimodal modeling as a specialized subset of the broader WFM framework, focusing our discussion within the complexities of cross-modal integration. Gu et al.~\citep{gu2025foundation} covers a comprehensive survey of unimodal WFMs.
% \yz{multimodal and longitudinal-aware training?}

% \textbf{Data Collection.}
\subsection{Data Acquisition}

\begin{table*}
% \vspace{-10pt}
\setlength{\tabcolsep}{5pt}
\renewcommand{\arraystretch}{1.1}
\caption{
\textbf{Overview of Multimodal Physiological Datasets Commonly Used for WFMs.}
The table summarizes commonly used datasets, their sensing modalities, temporal structure, scale, primary tasks, and availability. Scale is reported in number of recordings (rec), hours (h), nights (nts) or participants (pts).
% Contextual modalities represent non-sensor information such as demographic metadata or self-reported surveys. 
% \yw{let;s double check whether each dataset has contextual info}
% \yw{1) i believe every dataset have demographic info, should we include it? 2)shall we also mention task in this section or agentic part?..}\tx{I feel the contextual modalities are not informative. ECG-QA is not multi-modal, it only has ECG? And HAR-CoT only with ACC. How about TILEs?  Be aware of UKB, NAHANCE, and ALLofUS, somehow much larger, with multiple sensory data} \ds{rows 1 and 2 need to fill size/duration} 
% \ds{contextual modalities is not very clear as a column, let's explain it in the caption, do we mean metadata / non-sensor data?}
}
\vspace{-8pt}
\label{tab:dataset-overview}
\small
\begin{center}
\resizebox{\linewidth}{!}{
\begin{tabular}{l l c l l c}
\toprule[1.5pt]
\textbf{Dataset} & \textbf{Sensing Modalities} & \textbf{Longitudinal} & \textbf{Scale} & \textbf{Task} & \textbf{Open} \\
\midrule\midrule

Apple Heart and Movement Study~\citep{erturk2025beyond} &
PPG, ECG  &
% demographics, surveys &
\cmark &
2.5B h &
AFib, Fitness, Activity &
\xmark \\

Google Fitbit data~\citep{Narayanswamy2025lsm} &
ACC, PPG, EDA, Temp., Alt. &
% demographics &
\cmark &
40M h &
Activity, Metabolic health, Sleep &
\xmark \\

CLIMB~\citep{dai2025climb} &
ECG, 2D and 3D scans &
% demographics &
-- &
44 mixed datasets &
LOS, 48h IHM, Clinical &
\cmark \\

Cuffless BP~\citep{kachuee2017cuffless} &
PPG, ECG, BP &
% \xmark &
\xmark &
339 h &
BP estimation &
\cmark \\

PPG-Dalia~\citep{reiss2019ppgdalia} &
ECG, PPG, IMU, GSR &
% \xmark &
\xmark &
15pts / 71 h &
heart rate estimation &
\cmark \\

MAUS~\citep{beh2021maus} &
ECG, PPG, GSR &
% \xmark &
\xmark &
22 h &
stress detection, cognitive load &
\cmark \\

Mendeley-YAAD~\citep{dar2022yaad} &
ECG, GSR &
% \xmark &
\xmark &
25 pts / 5 h &
emotion &
\cmark \\

EPHNOGRAM~\citep{kazemnejad2021ephnogram} &
ECG, PCG &
% \xmark &
\xmark &
61 h &
simultaneous multi-modal analysis &
\cmark \\

BIDMC~\citep{pimentel2017bidmc} &
ECG, PPG &
% demographics &
\xmark &
14 h &
respiratory rate estimation &
\cmark \\

PhysioNet 2018 Challenge~\citep{ghassemi2018physionet} &
EEG, EMG, EOG, ECG &
% \xmark &
\xmark &
2K pts / 15K h &
Sleep staging, arousal &
\cmark \\

SleepEDF~\citep{kemp2000sleepedf} &
EEG, EOG, EMG &
% demographic &
2 nights &
197 nts &
Sleep staging &
\cmark \\

TUEG~\citep{iyad2016tueg} &
EEG, ECG, EOG, EMG &
% \xmark &
$\sim$1.56 sessions &
27K rec (EEG) &
pathology decoding from EEG &
\cmark \\

DEAP~\citep{koesltra2012deap} &
EEG, EMG &
% \xmark &
\xmark &
32 pts &
Emotion recognition &
\cmark \\

CAP~\citep{terzano2001cap} &
EEG, EOG, EMG &
% \xmark &
\xmark &
108 rec &
sleep stage, sleep disorders &
\cmark \\

Stanford Sleep Clinic (SSC)~\citep{kjaer2025stanford} &
EEG, EOG, EMG, ECG, Resp. &
% demographics &
partially multi-night &
35K rec / 280K h &
Sleep, 130+ diseases &
% ?\yw{?} \\
\cmark \\

SHHS~\citep{zhang2018shhs} &
ECG, respiratory (THX, ABD) &
% demographics &
follow-up in years &
6K rec, 46K h &
Sleep, CVD &
\cmark \\

MrOS~\citep{blank2005mros} &
ECG, respiratory (THX, ABD) &
% \xmark &
follow-up in years &
4K rec / 41K h &
Sleep, Aging &
\cmark \\

MESA~\citep{chen2015mesa} &
ECG, respiratory (THX, ABD), PPG &
% demographics, surveys, biomarker &
\xmark &
2K rec, 15K h &
Sleep, CVD, Metabolic &
\cmark \\

BioSerenity~\citep{thapa2026sleepfm} &
BAS, ECG, EMG, respiratory &
% demographics &
\xmark &
19K rec / 152K h &
N/A &
\xmark \\

% \midrule
WESAD~\citep{schmidt2018wesad} &
IMU, PPG, ECG, GSR &
% demographics, psychological surveys &
\xmark &
15 pts &
Stress detection &
\cmark \\

MPDB~\citep{tao2024mpdb} &
EEG, ECG, EMG &
\xmark &
% \xmark &
35 pts &
Driving behavior &
\cmark \\

PMData~\citep{Thambawita2020pmdata} &
vital signs from Fitbit&
% demographics, health history&
5 month&
16 pts &
Stress detection, Sleep quality &
\cmark \\

LifeSnaps~\citep{yfantidou2022lifesnaps} &
vital signs from Fitbit &
% demographics, EMA, surveys&
4 month&
71 pts &
Stress resilience, Sleep disorder &
\cmark \\

GLOBEM~\citep{xu2022globem} &
Smartphone metrics, vital signs from Fitbit &
% demographics, surveys&
Multiple years&
497 pts &
Depression and Anxiety detection &
\cmark \\

AWFB~\citep{fuller2020awfb} &
vital signs from smartwatches &
% \xmark&
104 hours&
49 pts &
Calorie burn, Activity classification &
\cmark \\

PHIA~\citep{merrill2026phia} &
vital signs from Fitbit &
% \xmark&
10 days&
3K pts &
Health insights questions answering &
\xmark \\

CoSt~\citep{hayat2023cost} &
Smartphone metrics &
% surveys&
16 weeks&
48 pts &
Risk of failing courses &
\xmark \\

PWUD~\citep{tyler2025pwud} &
Smartphone metrics &
% surveys&
30 days&
68 pts &
Risk of taking drugs &
\xmark \\

% HAR-CoT~\citep{} &
% ACC &
% \xmark&
% \xmark&
% -- &
% Human activity reasoning task &
% \cmark \\

Sleep-CoT~\citep{langer2025opentslm} &
EEG, EOG, EMG&
% \xmark&
2 nights&
197 nts &
Sleep stage reasoning task &
\cmark \\

% ECG-QA-CoT~\citep{} &
% ECG &
% \xmark&
% \xmark&
% -- &
% ECG-QA reasoning task &
% \cmark \\

\bottomrule[1.5pt]
\end{tabular}}
\end{center}
\vspace{-0.5cm}
\end{table*}

Wearable health monitoring includes continuous observation of physiological and behavioral signals using devices such as smartwatches, rings, ear-worn sensors, and smartphones~\citep{Lubitz2022DetectionOA, Liu2025EarMeterCR, Svensson2024ValidityAR}. Common wearable and mobile sensors include inertial measurement units (IMUs) for motion tracking, photoplethysmography (PPG) for cardiovascular monitoring, electrodermal activity (EDA) for sympathetic nervous system activity, temperature sensors, and electrophysiological recordings such as electroencephalography (EEG), electrocardiography (ECG), and electromyography (EMG). These modalities differ in temporal resolution, noise characteristics, and information they provide~\citep{gu2025foundation}.
% A more detailed introduction to these sensor modalities, including signal characteristics and typical use cases, is provided in Appendix~\ref{app:modalities}. \hj{Are we going to add the appendix? Then we can move the Data Diversity and Scale part to the appendix.}

We provide a summary of common datasets for training and evaluating multimodal WFMs in Table~\ref{tab:dataset-overview}.
Wearable data differs fundamentally from conventional healthcare and multimodal learning benchmarks in vision or language~\citep{Palmerini2023MobilityRB}. First, wearable data is \emph{multimodal}, combining heterogeneous sensor signals with different temporal resolutions, and contextual meaning. 
For example, WFMs such as PhysioOmni~\citep{jiang2025physioomni} are trained on datasets like SleepEDF~\citep{kemp2000sleepedf}, which incorporate EEG, EMG, EOG, and ECG signals. 
In addition to raw signals, some datasets also contain \emph{contextual} information,  higher-level or auxiliary information that situates raw signals within behavioral, environmental, or semantic context~\citep{abowd1999context, gu2025foundation}. For instance, GLOBEM~\citep{xu2022globem} and WESAD~\citep{schmidt2018wesad} include complementary self-reported emotion surveys alongside wearable signals.
Second, wearable signals are \emph{longitudinal by default}, as measurements are collected repeatedly over extended periods, capturing both short-term temporal dependencies and long-term evolution. This property shifts the modeling focus to within-subject dynamics, rather than population-level, single-point snapshots.

In terms of data scale, although wearable devices have the potential to generate continuous, long-term sensor streams, only a small number of existing WFMs currently operate at scales comparable to those seen in vision or language domains. Recent efforts such as LSM~\citep{Narayanswamy2025lsm} and LSM-2~\citep{xu2025lsm2}, trained on 40 million hours from 165K participants, and WBM~\citep{erturk2025beyond}, which uses 2.5 billion hours from 162K individuals, represent notable exceptions. Nevertheless, most wearable datasets, such as PPG-Dalia~\citep{reiss2019ppgdalia}, remain substantially smaller and fragmented across studies. 
% \yw{@Evelyn, what should we talk about UK biobank here?}

% These identified works vary in training scale, reflecting a transition from specialized clinical cohorts to massive longitudinal studies. Scale ranges from hundred-level hours in specialized tasks, such as Mathew et al. and UniCardio, to thousand-level hours in intermediate efforts like NormWear and BrainOmni that aggregate multiple public sources \qy{references for these work?}. The most recent frontier of biosignal foundation modeling reaches unprecedented scales, exemplified by LSM and LSM-2 training on 40 million hours from 165,000 participants, and WBM utilizing 2.5 billion hours from 162,000 individuals. 
% Data collection methods generally follow two paradigms: the strategic merging of existing public, synchronized multimodal datasets or the large-scale self-collection of longitudinal data via portable consumer devices. \hj{@Yvonne, please take a look on those paragraphs and check whether it is consistent with the prior one!}

\subsection{Modeling} 

Existing WFM modeling mainly focuses on training static modality-specific encoders. Early works put great effort into pretraining on large scales of unlabeled data to train effective encoders and have achieved strong classification/prediction on single modality tasks~\citep{xu2021limubert, pillai2025papagei, jiang2024labram}. Since wearable monitoring naturally involves a multimodal setting, in this section we mainly talk about how existing WFMs deal with multimodal alignment from different sensor-specific encoders. 
% \yz{still not convinced why we focus on multimodal...} 
Generally, existing multimodal pretraining for WFMs (as shown in Table~\ref{tab:fm-overview}) was mostly developed by following pipelines: i) \textit{Tokenization}, the process of encoding modality data into embeddings to extract features that represent the underlying physiological information; 
% \yw{we do not have this one in figure 1}
ii) \textit{Pretraining}, which defines the methodology for training the model on large-scale datasets, mainly using self-supervised objective such as contrastive or mask reconstruction, to learn generalizable representations; iii) \textit{Multimodal Alignment}, the approach for fusing heterogeneous embeddings from various modalities, allowing the model to process them in a unified representation space; iv) \textit{Integration of Language and Context}, which explores the use of LLMs to bridge biosignals with semantic insights and allows integration of diverse contextual information. 

\begin{table*}[t]
\setlength{\tabcolsep}{6pt}
\renewcommand{\arraystretch}{1.1}
\caption{
\textbf{Overview of Multimodal Biosignal Foundation Models.}
We summarize representative foundation models by their modality scope, tokenization strategy, multimodal alignment design, pretraining objective, and data scale. 
% \qy{I didn't find you refer to this table anywhere?}\ds{why the multimodal focus here?} \yz{ we need to explain the focus in 2.2 and ref this table}
}
\vspace{-8pt}
\label{tab:fm-overview}
\small
\begin{center}
\resizebox{\linewidth}{!}{
\begin{tabular}{l l l l l c}
\toprule[1.5pt]
\textbf{Model} & \textbf{Modalities} & \textbf{Tokenization} & \textbf{Alignment} & \textbf{Pretraining Objective} & \textbf{Scale} \\
\midrule\midrule

UniCardio &
ECG, PPG, BP &
Temporal patches &
Shared encoder &
Reconstruction &
$>300$ hrs \\

Fang et al. &
ECG, EEG, EOG, EMG &
Temporal patches &
Shared encoder &
Reconstruction &
$>15$K hrs \\

LSM &
HR, HRV, EDA, ACC, TEMP, ALT &
Numeric features &
Shared encoder &
Reconstruction &
$40$M hrs \\

LSM-2 &
HR, HRV, EDA, ACC, TEMP, ALT &
Numeric features &
Shared encoder &
Reconstruction &
$40$M hrs \\

NormWear &
ECG, PPG, EEG, GSR, IMU &
Spectral patches &
Shared encoder + CLS attention &
Reconstruction &
$>4$K hrs \\

Mathew et al. &
ECG, PPG &
Temporal and spectral patches &
Shared encoder + modality PE &
Reconstruction &
$>300$ hrs \\

BrainOmni &
EEG, MEG &
Temporal and spectral patches &
Shared encoder + modality PE&
Reconstruction &
$>2$K hrs \\

PhysioOmni &
ECG, EEG, EOG, EMG &
Temporal and spectral patches &
Shared + private encoders &
Reconstruction &
$>27$K hrs \\

FreqMAE &
MIC, GEO, ACC, GYR, MAG, LIGHT &
Spectral patches &
Shared + private encoders &
Reconstruction &
unspecified \\

bioFAME &
EEG, EOG &
Spectral patches &
Channel-independent encoder &
Reconstruction &
$>1$K hrs \\

PhysioWave &
ECG, EEG, EMG &
Spectral patches &
Late fusion &
Reconstruction &
$>700$ hrs \\

SleepFM &
EEG, EOG, EMG, ECG, RESP &
Temporal patches &
Shared encoder &
Contrastive learning &
$>585$K hrs \\

WBM &
Behavioral features &
Numeric features &
Shared encoder &
Contrastive learning &
$2.5$B hrs \\

\bottomrule[1.5pt]
\end{tabular}}
\end{center}
% \vspace{-0.6cm}
\end{table*}

\noindent \textbf{Tokenization.} Tokenization converts heterogeneous, high-frequency sensor streams into compact sequences of tokens via preprocessing and patch embedding, making them compatible with widely-known foundation model architectures such as Transformers~\citep{transformers} and Mamba~\citep{gu2024mamba}. A straightforward approach partitions sensor streams into fixed-length temporal windows and embeds each window into a patch token using a learnable encoder~\citep{chen2025unicardio,thapa2024sleepfm,thapa2026sleepfm,fang2024promoting}, which minimally alters the signal while preserving its temporal characteristics. Another line of works represents signals in the time–frequency domain~\citep{kara2024freqmae, luo2024normwear}, resulting in 2D patch tokens. %PhysioWave~\citep{chen2025physiowave} proposes a multi-layer wavelet tokenization to decompose signals into hierarchical frequency-band representations. 
Some works~\citep{xiao2025brainomni, liu2024frequencyaware} jointly leverage both time and frequency features, enabling their encoders to learn cross-domain representations. Models targeting longitudinal inputs employ statistical features to reduce token resolution: LSM~\citep{Narayanswamy2025lsm, xu2025lsm2} utilizes activity features (\textit{e.g.}, step counts) computed over 1-minute windows, while WBM~\citep{erturk2025beyond} uses features over hour-level windows. Modality-aware tokenization strategies are also used: Mathew et al.\citep{mathew2024foundation} apply spectral patching for PPG while patching ECG in the temporal domain, and PhysioOmni\citep{jiang2025physioomni} generates frequency tokens for EEG and EMG while using temporal tokens for ECG and EOG. 

\noindent \textbf{Pretraining.} 
After tokenizing raw sensor data into a representation space, the next step is to learn grounded patterns from large-scale unlabeled data that correlate with latent categorical or numerical semantics. Existing encoder development is mainly built for each single sensor modality following the self-supervised pretraining mechanism, including either i) reconstruction-based or ii) contrastive learning-based pretraining objectives. Reconstruction-based pretraining, typically instantiated through masked autoencoding, is the most popular approach. 
Existing works~\citep{mathew2024foundation, liu2024frequencyaware, fang2024promoting, xiao2025brainomni, chen2025unicardio} randomly mask a portion of tokens (from 50\% to 80\%) within each modality and train the model to reconstruct the tokens. For example, 
NormWear~\citep{luo2024normwear} and LSM~\citep{Narayanswamy2025lsm} mask reconstruction to temporal- and modality-level imputation by masking specific temporal segments or modalities.
On the other hand, several works adopt contrastive learning objectives to train encoders.
% \yw{is there any multimodal WFMs train each encoder in contrastive learning way, other wise we can put unimodality model her>?}. 
% SleepFM~\citep{thapa2024sleepfm,thapa2026sleepfm} perform cross-modal contrastive learning by treating time-synchronized modality pairs as positives, and further introduce leave-one-modality-out pairing where a single modality is contrasted against the averaged embeddings of the remaining modalities. \yw{i think sleepFM should be put into alignmment part?}
For example, WBM~\citep{erturk2025beyond} adopts contrastive pretraining, motivated by the inclusion of sparse but highly informative variables (\textit{e.g.}, VO$_2$max), and constructs positive pairs at the subject level.

\noindent \textbf{Multimodal Alignment.} After tokenization, WFMs align modality-specific representations within unified backbones. A common approach is to %UniCardio~\citep{chen2025unicardio}, LSM~\citep{Narayanswamy2025lsm}, LSM2~\citep{xu2025lsm2}, Fang et al.~\citep{fang2024promoting}, and WBM~\citep{erturk2025beyond} 
concatenate modality-specific tokens and feeds them into a shared encoder to learn joint representations~\citep{chen2025unicardio, Narayanswamy2025lsm, xu2025lsm2, fang2024promoting, erturk2025beyond, luo2024normwear}. %NormWear~\citep{luo2024normwear} adopts a similar design but applies self-attention specifically across the [CLS] tokens of each modality. 
Several works further incorporate modality-aware positional encoding to preserve modality %identity, including Mathew et al.~\citep{mathew2024foundation} and BrainOmni~\citep{xiao2025brainomni}. 
identity~\citep{mathew2024foundation, xiao2025brainomni}. PhysioOmni~\citep{jiang2025physioomni} and FreqMAE~\citep{kara2024freqmae} use both private modality encoders and a shared encoder operating on concatenated inputs, enabling the model to capture both modality-specific and cross-modal features. In contrast, PhysioWave~\citep{chen2025physiowave} adopts late fusion, where the outputs of modality-specific backbones are concatenated through a final layer for prediction. bioFAME~\citep{liu2024frequencyaware} enforces channel-independent encoding and employs multi-head filtering layers to produce consistent representations across heterogeneous channels, thereby facilitating multimodal alignment.
SleepFM~\citep{thapa2024sleepfm,thapa2026sleepfm} perform cross-modal contrastive learning by treating time-synchronized modality pairs as positives, and further introduce leave-one-modality-out pairing where a single modality is contrasted against the averaged embeddings of the remaining modalities. 
% \yw{i think sleepFM should be put into alignmment part?}

\noindent \textbf{Integration of Language and Context.} 
% Bridging Biosignals with Language and Context}
% Integration of Language and Contextual Reasoning
% As summarized in Figure xx, existing approaches can be broadly grouped into three paradigms: \textit{(i) text-centric encoding of time series}, \textit{(ii) alignment of biosignals with large language models (LLMs)}, and \textit{(iii) language-mediated question answering or reasoning over health data}.
Recent work has begun to incorporate language and contextual information into multimodal health modeling, aiming to bridge low-level biosignals with high-level semantic reasoning. 
Early efforts reformulate physiological signals as textual representations to directly leverage pretrained LLMs~\citep{pmlr-v248-kim24b, liu2023large}. While simple and flexible, this paradigm heavily depends on handcrafted summarization schemes, resulting in information loss and limited scalability to fine-grained temporal modeling. 
% \qy{Are these modality alignment methods as well?}% To preserve signal integrity, more advanced architectures focus on modality alignment. 
Models such as SensorLLM~\citep{li2025sensorllm}, 
% LLaSA~\citep{asif2025llasa},
MedTSLLM~\citep{chan2024medtsllm},
MEIT~\citep{wan2025meit}, and RespLLM~\citep{pmlr-v259-zhang25a} adopt projection layers to map biosignals (\textit{e.g.}, IMU, ECG, audio) into the LLM embedding space. This facilitates sensor-conditioned language generation, such as report drafting or answering queries that incorporate contextual metadata like patient history and symptoms. In contrast, SensorLM~\citep{zhang2025sensorlm} utilizes a dual-encoder and multimodal decoder architecture. By optimizing via contrastive and captioning losses, it directly aligns sensor and text modalities, enabling zero-shot retrieval alongside standard generation.
These models demonstrate strong zero-shot or few-shot performance on tasks such as report generation, question answering (QA), and disease classification, highlighting the promise of language as a unifying fusion interface across heterogeneous modalities. 

% Building upon the alignment efforts, the most recent shift treats language as a medium for active reasoning and sensemaking. For example, OpenTSLM~\citep{langer2025opentslm} introduces chain-of-thought datasets to ground reasoning in temporal dynamics for tasks such as HAR and ECG QA. \yw{this one should move to inference?}
% , while emerging agent systems such as PHIA and GLOSS push toward open-ended insights and sensemaking.
% Despite these advances, a few gaps remain. \qy{If you want to point out the gaps, you'd better do the same thing on other subsections as well.}
% \yz{let's not add challenge in section 2}
% \yw{i commented the challenge}
% \begin{itemize}
%     \item Limited flexibility and scalability across different number of sensor modalities, which hinders the seamless integration of diverse, heterogeneous data streams without significant task and modality-specific tuning.
%     \item Insufficient longitudinal and personalized grounding, which fails to capture evolving physiological baselines and long-term health trajectories beyond short temporal windows.
% \end{itemize}

\subsection{Inference}
% \textbf{Inference and adaptation.} Pretrained models are adapted for downstream tasks such as health monitoring, risk assessment, or decision support. Inference may involve personalization, temporal reasoning, and integration of new observations over time.

Despite advances in pretraining scale and representation learning, most existing WFMs remain deployed as \textit{discriminative predictors} to specific patterns.
Downstream inference applications typically are task-specific classification or regression. 
% raising a fundamental question about what ``foundation'' means at the level of application, where models are expected to support flexible and general-purpose inference. 
For example, WBM~\citep{erturk2025beyond} is trained on over 2.5B hours of behavioral data, yet its downstream applications are limited to fixed prediction tasks such as sleep duration estimation, diabetes detection, or pregnancy detection through supervised fine-tuning. Similarly, LSM~\citep{Narayanswamy2025lsm}, NormWear~\citep{luo2024normwear}, and PhysioOmni~\citep{jiang2025physioomni} pretrain multimodal representations but evaluate them on predefined classification or regression benchmarks.

% Building upon the alignment efforts, the most recent shift treats language as a medium for active reasoning and sensemaking. For example, OpenTSLM~\citep{langer2025opentslm} introduces chain-of-thought datasets to ground reasoning in temporal dynamics for tasks such as HAR and ECG QA.

Recent language-aligned models take important steps beyond purely discriminative prediction but still largely remain bounded to closed-world formulations, including activity recognition with fixed label sets or question answering derived from predefined templates. % . By conditioning inference on text prompts, these models enable flexible task specification and support generative outputs. 
For example, SensorLM~\citep{zhang2025sensorlm} is capable of generating captions grounded in statistical, structural and semantic features. OpenTSLM~\citep{langer2025opentslm} utilizes question-answering pairs, enabling models to produce rationales alongside predictions. However, the scope of these experiments remains constrained when compared to contemporary LLMs and vision–language models. 
\section{Call to Action}
% This section outlines three key challenges of existing patter recognition centric WFMs from previous work and outline foundational shift to support more anticipatory and human-aligned health monitoring.
This section outlines the key challenges in current WFMs and proposes a foundational shift toward more anticipatory and human-aligned health monitoring. For clarity, we categorize these issues into data, modeling, and inference levels, though many challenges are intrinsically linked across categories.

% \subsection{Challenges:}
\subsection{Structurally Rich Data}
\noindent \textbf{Challenges.} Despite rapid progress in wearable sensing and the growing availability of multimodal physiological datasets (Table~\ref{tab:dataset-overview}), data remains a fundamental bottleneck for enabling WFMs to move beyond static encoders. The bottleneck lies not in volume, but in openness, unification of multiple heterogeneous datasets, and rich temporal and contextual information integration.

%1lack of open-source --> call for opensource or build large dataset?
First, most population-scale wearable datasets with millions of hours of longitudinal data remain proprietary, primarily originating from large industrial studies~\citep{erturk2025beyond, zhang2025sensorlm, xu2025lsm2}. While these datasets have enabled impressive demonstrations of predictive performance and range of application, their inaccessibility limits reproducibility, cross-dataset evaluation, and community-driven progress. In contrast, open-source datasets are typically orders of magnitude smaller, short-term, or restricted to narrow sensing modalities, and are often curated around specific downstream outcomes or tasks~\citep{reiss2012pamap2, schmidt2018wesad}.
% This suggest a significant gap towards truly open and generalizable wearable foundation models.
%2 unlike other domains, others datasets contain different resolutions,,need unifies method to training them into unified-->suggestion like UNITS? or other directions?///

Second, unlike vision or language domains, wearable datasets are inherently heterogeneous, which brings challenges in unification of existing datasets. 
% \hj{Can we say this as a challenge? I think it is more as a characteristic, we need to focus more on what is the resulting challenging from this heterogeneity. And if it is a challenge, for me, it sounds as a problem that is impossible to solve.} \hj{One other thing. While we say heterogeneity as a challenge we also note it as a future direction (data collection should prioritize heterogeneity).} 
Datasets in Table~\ref{tab:dataset-overview} vary diversely in sensor types, various sampling rates, temporal resolution, and recording conditions. As a result, current WFM studies often result in dataset-specific modeling and evaluation, (\textit{e.g.}, large ECG-only based or IMU-only based benchmarks~\citep{reiss2019ppgdalia}). 
% This fragmentation further entrenches pattern recognition by constraining models to narrow, dataset-bound input distributions.
% to handle fragmented datasets, resulting in siloed foundation models that fail to generalize across sources. 
% out-come based data collection/ here shall we talk about task or basic integrate contexual information?> --> we can curate new datasets using contextual information / llm, or rag help??

Moreover, most wearable datasets focus only on raw and short-term physiological signals and behavioral metrics.  As a result, they fail to capture long-term patterns, spanning multiple days, weeks or even years, that reflect noticeable changes and different states within the same individual. The absence of such continuous, long-term data restricts models from learning personalized trajectories and tracking changes over time, which are critical for health understanding and cannot be captured through fixed temporal windows alone.

% The lack of long-term continuous data restricts models from learning personalized trajectories and tracking changes over time, which are key signals for health understanding that cannot be captured through fixed temporal windows alone.\hj{Regarding this longitudinal data, it would be better to be specific. How long should it be? - Like week or month. At the same time, I think there are few datasets that has 24-hour monitoring which we can say as long-term data.}
Finally, limited attention to complementary contextual information, such as demographics, existing medical conditions~\citep{abowd1999context}, or insights into what the person is experiencing in real time, further constrains model learning. 
% Consequently, models trained on these datasets are typically outcome-conditioned, capturing what happened but not why it happened, which limits their ability to support personalized health reasoning or planning tasks. 
% \hj{Maybe we can remove the prior parts and simply focus on this point (data should be paired with context). This is new to the community, and actionable.} 

% \subsection{Recommendation: Toward open, heterogeneous and scalable data ecosystem}
\noindent \textbf{Recommendation: Toward open, context-aware and scalable structurally rich data ecosystem.} 
% \hj{context is not included in the title here.} 
% \yz{the third word should be scalable unification of the heterogeneous / fragmented datasets? What's the best word..}
We argue that existing data construction practices for WFMs structurally reinforce a static encoder paradigm, and therefore require a fundamental rethinking of the data ecosystem. Rather than treating data scale alone as the primary driver of progress, future research must prioritize the availability of structurally rich data, that integrates multimodal sensor streams, long-term personal trajectories, and contextual metadata within a unified framework. This requires moving beyond isolated datasets and outcome-conditioned collection toward open and interoperable data ecosystems that support longitudinal continuity, cross-dataset reuse, and semantic grounding.

% \emph{openness}, semantic context, and unification of hetergeneity as first-class design objectives, advancing them in a coordinated manner across data collection, data organization, and representation alignment.
% \yw{also, the data richness term is not here yet..to help move beyond fixed pattern aquisition?}.

To achieve this, first from a data collection perspective, we call for greater investment in long-term, large-scale, and openly accessible wearable datasets to complement closed large-scale industrial studies. For example, the UK Biobank~\citep{sudlow2015uk} and All-of-US~\citep{rutter2019all} studies provide a few valuable resources (though under limited access) that enables population-scale analysis pairing wearable data with clinical context over extended longitudinal periods. Future efforts should build on these initiatives by promoting openness and ease of access, expanding the range of collected modalities, increasing the coverage of wearable data across larger portions of the population and across follow-up periods, and ultimately making such datasets more suitable for WFMs training.

At the same time, meaningful progress does not necessarily depend on collecting new data. Instead, the community should prioritize benchmark construction by developing standardized preprocessing tools that enable joint training across datasets, mitigate fragmentation, and accommodate sensor diversity without bias toward dataset- or device-specific signal patterns. %and alleviate the fragmentation across various datasets and include sensor diversities and avoid bias to specific sensor pattern\jh{3 "and" in this sentence, rewrite to make it clearer?}.
%to avoid homogeneity to bias to specific data pattern?
% Even with careful benchmarking, perfect alignment across datasets is often infeasible due to fundamental differences in sensing resolution or temporal granularity. Therefore, representation-level alignment and fusion beyond raw signal level can be promising direction. 
Another alternative lies in modeling: instead of forcing raw data harmonization, future WFMs should be trained under shared representation frameworks that map heterogeneous inputs into common latent spaces, enabling partial observability and cross-dataset transfer.

\subsection{Multimodal and Longitudinal Modeling}

\textbf{Challenges.} Current %Foundation Models for Wearables (WFMs) 
WFMs are largely constrained by modeling paradigms borrowed from computer vision and natural language processing, which are often ill-suited for the nuances of physiological data. Most existing approaches remain focused on single modalities or fixed, task-specific combinations. Early efforts to build multimodal WFMs exist (Table~\ref{tab:fm-overview}), but lack standardized protocols (\textit{e.g.}, for tokenization and architectural fusion), so it remains difficult to compare performance across studies or generalize these models to new, heterogeneous sensor sets. This fragmentation forces the field to remain dataset-driven rather than developing truly universal representations.

This static encoder paradigm is further compounded by a widespread failure to integrate complementary contextual information, such as demographics, medical history, or real-time environmental insights, during the pretraining process. By ignoring these semantic anchors, they may accurately capture a physiological event, such as a heart rate spike, but they lack the causal context to distinguish between a chronic medical condition and an acute environmental stressor. Consequently, without this contextual grounding, WFMs capture what occurred in the sensors but struggle to explain what happened with the human, which severely limits their utility for personalized health reasoning or long-term clinical planning.

Perhaps most critically, despite naturally operating on a unique form of data that are collected and changing continuously every day, WFMs remain predominantly short-horizon targeting static predictive tasks, lacking attention and exploration for longitudinal modeling. The common practice of processing data as isolated, cross-sectional windows disrupts the temporal continuity needed to track health trajectories. Transitioning toward longitudinal-aware modeling, these models need to balance high-frequency physiological sampling with the computational efficiency needed to represent the data, taking into consideration the multi-scale nature of physiological change. We also need to consider the inherent sparsity and irregularity of real-world data, where modalities frequently appear, disappear, or degrade due to device changes and user behavior, which would be exacerbated over longer timespans.

% Transitioning toward longitudinal-aware modeling, we 
% Barriers in extending to longitudinal-aware modeling include balancing high sampling rates of physiological signals with representation efficiency over longer periods, as well as dealing with the prominent missingness and irregularity of modalities. 

\textbf{Recommendation: Toward cross-modal and longitudinal modeling pipeline.}
% \yz{wording for any number and any combination? modality-agnostic? omni-modal?}\hj{I updated the paragraph. And how about ``cross-modal''? It better reflects the transferability.}
To move beyond modality-specific progress, the community should prioritize systemic generalization, where modeling insights transfer across studies through \textit{functional interoperability}. %, ensuring different systems can work together and produce comparable results even if their internal architectures differ. 
% Specifically, models should be able to process the same input types and produce standardized outputs. We should arrive at a unified protocol to make models truly comparable and also adaptable to new, heterogeneous datasets. 
Rather than pursuing a single universal architecture, progress should be organized around shared evaluation interfaces that allow independently developed models to be meaningfully compared, reused, and extended. 
Similar to leaderboards in NLP or CV, we recommend establishing common benchmarks for biosignals. By evaluating models on popular modality subsets (\textit{e.g.}, IMU-only, EOG+EEG, or PPG+ECG) within open datasets, the community can perform rigorous, ``apples-to-apples'' comparisons of model effectiveness.
This approach respects the natural diversity of biosignal sensors while providing a shared evaluative anchor. It ensures
modeling insights transfer across studies and reduces the need for exhaustive re-implementation  adapting to new sensor configurations.
%that WFMs can eventually function as interoperable components within a larger agentic ecosystem.

Beyond raw sensor data, we propose the deep integration of rich, multi-scale \textit{contextual information} during the pretraining phase. Even when demographics, medical histories, or environmental insights are sparse and unstructured, they provide the necessary semantic grounding to move models beyond simple outcome-conditioning. One promising path forward involves LLMs through sophisticated alignment and fusion techniques. By treating context as a primary input rather than a secondary metadata field, WFMs can learn to interpret physiological signals through a clinical or behavioral lens, enabling the transition from ``what'' happened in sensor readings to a personalized and grounded understanding of ``why'' it occurred.

Finally, future WFMs must shift from static, cross-sectional windows toward longitudinally-aware architectures capable of modeling health trajectories over months or years instead of seconds. This requires moving beyond simple extension of sequence length, but to adopt hierarchical temporal representations that compress high-frequency streams into coarser, multi-scale tokens (\textit{e.g.}, from heartbeats to weekly behavioral patterns). We argue for training objectives grounded in trajectory forecasting~\citep{Aminikhanghahi2016ASO} and event discovery~\citep{Dang2023ConditionalNO}, similar to blueprints found in EHR and medical imaging~\citep{Lan2021IntraInterSS,Raghu2023SequentialMS}. To handle the inherent sparsity of real-world sensing, longitudinal modeling should be reframed as selective evidence retrieval. Inspired by retrieval-augmented generation (RAG), WFMs can maintain \textit{personal memory} as a persistent, queryable health history that remains robust even when specific modalities are missing or asynchronous.

\subsection{The Need for Agentic Inference}

\noindent \textbf{Challenges.} 
Finally, at the inference stage, %where WFMs are deployed to specific applications
existing WFMs mainly function as \textit{discriminative predictors}, implicitly framing health monitoring as classification or regression tasks. Specifically, existing discriminative WFMs answer the question: ``Given the last 10 minutes of data, what is the label?'' 
% \yw{let's add hidden pattern-recognition training/prediction manner?} \hj{I changed the sentence.}
However, real-world health use cases require addressing questions like: ``What should I attend to next?'', ``Is there sufficient evidence to draw a conclusion?'', and ``How should my behavior change as new observations arrive?'' 
% \jh{? behaviour? not sure what belief is in this context, belief of my health status} 
%These questions cannot be resolved by one-shot prediction alone. Addressing them requires not only aggregating information across modalities and time, but also deciding how inference should proceed as evidence evolves. 
In other words, practical health requires models with \textit{agentic inference capability} to iteratively reason over multimodal, longitudinal signals by selecting what to infer, when to infer it, and how to update its internal state as evidence evolves, providing control over the inference process itself. Existing WFMs provide the representational capacity for health AI to capture pattern and correlation, but their capability for such proactive agentic inference remains largely unexplored.
% In real-world settings, health cannot be reduced to a single prediction problem; it unfolds as a sequence of decisions over time. 
% This shifts the problem from static pattern recognition prediction to adaptive control over the inference process.

% \textit{Agentic inference} refers to the ability of a model to iteratively reason over multimodal, longitudinal signals by selecting what to infer, when to infer it, and how to update its internal state as evidence evolves, providing control over the inference process itself. 
% Foundation models provide the representational capacity for health AI, but without agentic inference, their usage remains confined to task-specific prediction. We argue that this agentic inference is needed for real-world health tasks.

While emerging research~\citep{heydari2025pha, merrill2026phia, yang2025contextagent, choube2025gloss} has explored agentic frameworks for health (\textit{e.g.,} personal health coaching and context-aware services), a significant gap remains in their \textit{interpretation-level depth}. Existing agents often rely on shallow abstractions, such as static feature extractors or basic code generation, rather than leveraging the high-dimensional physiological representations learned by modern WFMs. Consequently, there is an architectural disconnect between agentic frameworks and WFMs: foundation models provide the necessary representational capacity, but current agentic systems lack the deep integration required to utilize these models as sophisticated interpretive tools. This lack of a proactive, signal-grounded inference layer prevents health AI from evolving beyond passive observation into adaptive, closed-loop systems.

\noindent \textbf{Recommendation: WFMs as a Bio-Toolset.} 
To address this gap, we recommend organizing WFMs as a coherent \textit{Bio-Toolset}: a collection of modular components that can be readily integrated into agentic pipelines. In this view, WFMs are not treated as end-to-end health decision-makers, but as high-fidelity perceptual and analytical components that provide actionable intelligence to a centralized reasoning agent. %This re-framing (i.e., from ``models as solutions'' to ``models as tools'') is motivated by three core requirements of agentic health systems.
%At the same time, this shift
This re-framing (\textit{i.e.}, from ``models as solutions'' to ``models as tools'') provides the interpretation-level depth that existing agentic frameworks currently lack by replacing shallow code-based heuristics with high-dimensional physiological representations. This modularity enables the system to handle the compositional complexity of real-world health, which requires multi-step verification, such as calling a denoising expert to validate signal integrity before invoking a diagnostic expert for event localization and a longitudinal retriever for baseline contextualization. %Furthermore, connecting agents with WFMs allows the system to move away from black-box outputs toward reasoning grounded in verifiable evidence, while supporting dynamic resource allocation by selectively invoking lightweight experts for routine monitoring and reserving high-capacity models for complex anomalies. This architectural decoupling of sensing from reasoning ensures that health AI remains transparent, computationally efficient, and adaptable to the evolving landscape of wearable sensors.
Through this transition, we shift the role of WFMs from static discriminative predictors focused on specific patterns toward foundational components of an agentic system. By decoupling physiological sensing from clinical reasoning as distinct functional roles, while tightly coupling them through an agentic interface, we can enable open-ended, adaptive inference that evolves alongside the user's health trajectory. 
\section{Alternative Views}
\textbf{A sufficiently large single-modality foundation model is enough.}
Some researchers argue that the field should focus on training very large foundation models on a single, information-rich modality (\textit{e.g.}, PPG, ECG, or IMU). Proponents of this view point to recent successes where large unimodal models achieve strong performance across multiple downstream tasks, including sleep staging, arrhythmia detection, and activity recognition. For example, studies on single-modality foundation models~\citep{jiang2024labram, pillai2025papagei} demonstrate that, with sufficient data scale and carefully designed pretraining objectives, unimodal representations can generalize across populations and tasks. \\
\textbf{Response:} We agree that large single-modality foundation models provide a strong and often necessary baseline, particularly when multimodal data are unavailable or noisy. However, unimodal scaling alone fundamentally limits the scope of health understanding achievable from wearable data. Many clinically and behaviorally meaningful states, such as stress, fatigue, medication effects, or lifestyle changes, manifest through interactions between physiological, behavioral, and contextual signals rather than within a single modality. Moreover, unimodal models remain constrained to static encoders within that specific modality and struggle to support cross-signal reasoning. As a result, unimodal WFMs are well-suited for detection tasks but insufficient for comprehensive, longitudinal health reasoning.

%our response
\textbf{Prediction is sufficient; intervention and decision-making should remain human-led.}
Some research holds that healthcare systems including WFMs should be limited to prediction and monitoring, while interpretation, diagnosis, and intervention should remain strictly human-led~\citep{whicher2023artificial, Lekadire081554}. This view is motivated by legitimate concerns around safety, regulation, accountability, and ethical risk, particularly in sensitive domains such as mental health or clinical decision-making. \\
\textbf{Response:} We agree that WFMs must augment, not replace, clinical judgment. Our position does not advocate for purely autonomous decision-making, but for \textit{agentic support} that augments human decision-making while remaining firmly human-in-the-loop~\citep{gondocs2024aiprediction}.
In this role, agentic WFMs function as supportive reasoning systems that assist with longitudinal planning and evidence aggregation. For example, such systems can support adaptive monitoring by deciding which signals to summarize, over what temporal window, and how new observations relate to an individual’s historical baseline prior to a clinical visit. %This shifts the model’s role from prediction to contextual sensemaking.
Crucially, this agentic inference can alleviate the mounting pressure on national healthcare systems by providing patients with more accessible, flexible tools for proactive health management.
% By shifting the model’s role from raw prediction to contextual sensemaking, we empower both clinicians and users with actionable insights rather than just more data.

% \dm{perhaps add a sentence to highlight the benefit of agentic inference, for example, alleviate the pressing challenges for national healthcare systems, offering better accessibility and flexibility for users/patients in managing their health, etc...}
\section{Conclusion}

This paper calls for a fundamental shift in how the next generation of wearable foundation models should move beyond static encoders and be explicitly designed for longitudinal, anticipatory health reasoning. 
% Specifically, three fundamental shifts are required to enable this transition:
% \textbf{(i)Structurally Rich Data:} Moving beyond isolated datasets and outcome-conditioned collection toward open, interoperable data ecosystems that integrate multimodal sensing, long-term personal trajectories, and contextual metadata. \textbf{(ii)Longitudinal-aware Multimodal Modeling:} Prioritizing long-context inference, temporal abstraction, and personalization over cross-sectional or population-level prediction to enable reasoning over evolving personal health history. \textbf{(iii)Agentic Inference Systems:} Moving beyond static, retrospective prediction to support adaptive, human-aligned health support, including planning and decision-making under uncertainty.
The proposed direction serves as a starting point for rethinking the building of the data ecosystem, longitudinal-aware multimodal modeling, and agentic inference to drive the anticipatory health support for future WFMs. We argue that adopting these directions will enable wearable foundation models to evolve into longitudinal, context-aware, and clinically grounded foundations that can meaningfully support continuous and accessible health monitoring.
% in everyday settings.

\newpage
\section*{Acknowledgements}
This work was supported by Nokia Bell Labs through a donation and EPSRC grant EP/Z53447X/1. and  Y. Z. is additionally supported by the Cambridge Trust Scholarship.

\bibliography{example_paper}
\bibliographystyle{icml2026}

%%%%%%%%%%%%%%%%%%%%%%%%%%%%%%%%%%%%%%%%%%%%%%%%%%%%%%%%%%%%%%%%%%%%%%%%%%%%%%%
%%%%%%%%%%%%%%%%%%%%%%%%%%%%%%%%%%%%%%%%%%%%%%%%%%%%%%%%%%%%%%%%%%%%%%%%%%%%%%%
% APPENDIX
%%%%%%%%%%%%%%%%%%%%%%%%%%%%%%%%%%%%%%%%%%%%%%%%%%%%%%%%%%%%%%%%%%%%%%%%%%%%%%%
%%%%%%%%%%%%%%%%%%%%%%%%%%%%%%%%%%%%%%%%%%%%%%%%%%%%%%%%%%%%%%%%%%%%%%%%%%%%%%%
% \newpage
% \appendix
% \onecolumn
% \section{You \emph{can} have an appendix here.}

% You can have as much text here as you want. The main body must be at most $8$
% pages long. For the final version, one more page can be added. If you want, you
% can use an appendix like this one.

% The $\mathtt{\backslash onecolumn}$ command above can be kept in place if you
% prefer a one-column appendix, or can be removed if you prefer a two-column
% appendix.  Apart from this possible change, the style (font size, spacing,
% margins, page numbering, etc.) should be kept the same as the main body.
% \input{sections/06_appendix}
%%%%%%%%%%%%%%%%%%%%%%%%%%%%%%%%%%%%%%%%%%%%%%%%%%%%%%%%%%%%%%%%%%%%%%%%%%%%%%%
%%%%%%%%%%%%%%%%%%%%%%%%%%%%%%%%%%%%%%%%%%%%%%%%%%%%%%%%%%%%%%%%%%%%%%%%%%%%%%%

\end{document}